\documentclass[final,5p,times,twocolumn]{elsarticle}

\journal{Pattern Recognition Letters}

\usepackage{graphicx}
\usepackage{multirow}
\usepackage[hidelinks]{hyperref}
\usepackage{xcolor}
\newcommand{\mstd}[2]{\ensuremath{#1\,{\scriptstyle\pm}\,#2}}
\usepackage{booktabs}
\usepackage{algorithm}
\usepackage{algpseudocode}
\usepackage{amsmath}
\DeclareMathOperator*{\argmax}{arg\,max}
\usepackage{amssymb}

\begin{document}

\begin{frontmatter}

\title{Lag-aware cross-hand alignment for dual-hand action segmentation}

\author[ut]{Fatemeh Ziaeetabar\corref{cor1}}
\ead{fziaeetabar@ut.ac.ir}

\cortext[cor1]{Corresponding author.}
\address[ut]{Department of Computer Science, School of Mathematics, Statistics and Computer Science, College of Science, University of Tehran, Tehran, Iran}

\begin{abstract}
Dual-hand action segmentation commonly fuses left- and right-hand representations at identical temporal indices, although coordinated hand transitions may occur with nonzero and time-varying delays. We introduce Lag-Aware Cross-Hand Alignment (LACA), a lightweight module that explicitly estimates directional temporal-offset distributions between hand-specific feature streams. LACA retrieves cross-hand information from the estimated offsets and incorporates a learned null state to suppress transfer when no compatible cross-hand transition is supported. Alignment is supervised using compatibility-aware targets derived automatically from frame-level training annotations, without requiring additional labels. Analysis of the HA-ViD and ATTACH training annotations reveals robust nonzero cross-hand matches for 44.7\% and 48.9\% of transition anchors, respectively, compared with 18.6\% and 21.3\% under temporally shifted controls. When integrated into Polyphony, LACA improves the two-hand mean F1@50 from 40.4 to 42.5 and boundary F1 from 56.5 to 59.6 on HA-ViD, and from 19.9 to 21.8 and 44.7 to 47.9, respectively, on ATTACH, relative to our reproduced Polyphony baseline. These gains require only approximately 0.0086 million additional trainable parameters. We further introduce LACA-C, a future-free variant that restricts alignment and the complete inference pipeline to current and past observations. On ATTACH, LACA-C achieves 83.6\% transition-cue recall, a seed-averaged median availability delay of 233~ms, 0.72 false cues per minute, and segmentation-stage throughput of 224.9 current-position predictions per second. These results demonstrate that explicit cross-hand temporal alignment improves both action segmentation and boundary localization while supporting timely future-free perception.
\end{abstract}

\begin{keyword}
Dual-hand action segmentation \sep bimanual interaction \sep
temporal alignment \sep cross-hand coordination \sep
streaming action segmentation
\end{keyword}

\end{frontmatter}


\section{Introduction}
\label{sec:introduction}

\begin{figure*}[t]
\centering
\includegraphics[width=\textwidth]
{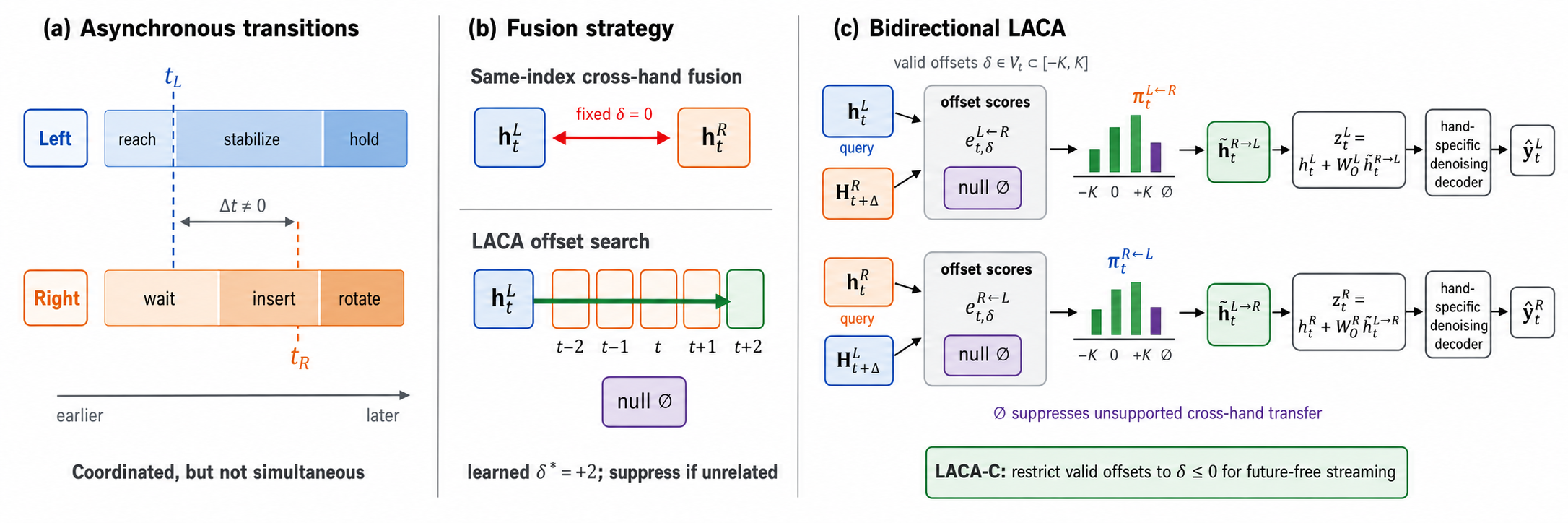}
\caption{Motivation and overview of lag-aware cross-hand alignment.
(a) Left- and right-hand label timelines in an assembly sequence, where the left-hand transition at $t_L$ precedes the corresponding right-hand transition at $t_R$, illustrating a nonzero temporal lag.
(b) Conventional same-index cross-hand fusion exchanges features using the fixed offset $\delta=0$, whereas LACA searches over valid nearby temporal offsets and suppresses cross-hand transfer through the null state when no compatible relation is supported.
(c) For each alignment direction, $L \leftarrow R$ and $R \leftarrow L$, LACA jointly scores the target-hand query, candidate opposite-hand features, and an explicit null state; estimates a directional offset distribution; retrieves an aligned cross-hand representation; and applies a residual update that conditions the corresponding hand-specific denoising decoder. LACA-C restricts valid offsets to $\delta \leq 0$ within a future-free, prefix-wise inference pipeline. Arrows indicate temporal information-retrieval and alignment directions rather than causal influence between the hands.}
\label{fig:laca_overview}
\end{figure*}

Consider an assembly operation in which one hand stabilizes a component while the other subsequently inserts or rotates a second part. Although the hands cooperate, their action transitions need not occur simultaneously: the stabilizing hand may enter a holding phase before the manipulating hand begins insertion, and the two hands may complete their actions at different times (Fig.~\ref{fig:laca_overview}(a)). Dual-hand action segmentation must therefore determine not only what each hand is doing, but also when its state changes. Accurate hand-specific temporal boundaries are important for fine-grained activity understanding, particularly in online settings where early or delayed transition estimates may reduce the usefulness of downstream cues. Such information may also support time-sensitive applications, including robot-assisted assembly and human--robot collaboration \cite{matheson2019humanrobot,aganian2023attach}.

%

Most dual-stream architectures fuse left- and right-hand features at corresponding temporal indices. Polyphony~\cite{zheng2026polyphony}, the closest recent dual-hand segmentation framework, also uses same-index cross-hand fusion and therefore leaves the temporal offset between related transitions implicit. Near asynchronous boundaries, this may combine features from different semantic phases, as illustrated in Fig.~\ref{fig:laca_overview}(b).

We introduce \emph{Lag-Aware Cross-Hand Alignment} (LACA), a lightweight module that retrieves cross-hand information using explicit bidirectional distributions over candidate temporal offsets. At each time step, LACA estimates separate alignment distributions for the $L\leftarrow R$ and $R\leftarrow L$ directions and uses them to construct temporally aligned cross-hand representations. An additional null state allows the model to suppress information transfer when no compatible cross-hand event is present. Alignment learning is guided by compatibility-aware transition targets
constructed automatically from existing frame-level labels: transition
types are matched using statistics computed exclusively from the training
split, while unmatched or incompatible target-hand transition anchors are
assigned to the null state. We further introduce LACA-C, a future-free variant that restricts alignment to current and preceding frames for future-free streaming analysis. Importantly, alignment direction represents temporal information retrieval rather than causal influence between the hands.

Our contributions are threefold:
\begin{enumerate}
\item We quantify nonzero cross-hand transition lags in frame-level annotations and compare them with temporally shifted controls.

\item We introduce a lightweight lag-aware alignment module that estimates time-varying cross-hand offsets and suppresses unsupported feature transfer.

\item We evaluate LACA on HA-ViD \cite{zheng2023havid} and ATTACH \cite{aganian2023attach} using segmentation and boundary-localization metrics, and assess its future-free variant using only current and past observations.
\end{enumerate}

Together, these link asynchronous bimanual transitions with explicit cross-hand alignment and future-free segmentation.

\section{Related Work}
\label{sec:related_work}

Hand-resolved assembly datasets such as HA-ViD \cite{zheng2023havid} and ATTACH \cite{aganian2023attach} have enabled fine-grained modeling of left- and right-hand actions. DuHa \cite{zheng2024duha} combines hand--object graph features, scene-level representations, and temporal modeling, whereas DuCAS \cite{zheng2024ducas} decomposes actions into verb, manipulated-object, target-object, and tool components followed by knowledge-based refinement. Although both methods address dual-hand segmentation, they require ground-truth object bounding boxes and do not explicitly estimate time-varying offsets between hand-specific transitions. Graph-based approaches have also modeled bimanual manipulation through hierarchical scene graphs, and multimodal relational reasoning \cite{ziaeetabar2024hierarchical,ziaeetabar2025adaptive, ziaeetabar2025efficientgformer}. While effective for representing hand--object and object--object relations, these formulations leave temporal dependencies between the two hands largely implicit and do not explicitly align asynchronously occurring hand transitions.

Polyphony \cite{zheng2026polyphony} combines alternating representation learning, semantic conditioning, diffusion-based prediction, and asymmetric inter-hand fusion. However, its cross-hand interaction is performed at corresponding temporal indices, leaving the offset between related transitions unmodeled. Related temporal methods address other forms of temporal variability: OnlineTAS \cite{zhong2024onlinetas} uses attention-based context augmentation and adaptive memory for future-free segmentation, TDRN \cite{lei2018tdrn} learns deformable temporal sampling offsets, and D$^3$TW \cite{chang2019d3tw} establishes global correspondences between weakly aligned sequences. These methods improve temporal context aggregation or sequence alignment, but do not explicitly identify which temporally offset transition of one hand is associated with the current state of the other.

LACA addresses this gap by learning explicit, directional, and time-varying cross-hand lag distributions, together with a null state that suppresses information transfer when no compatible cross-hand transition is supported.

\section{Proposed Method}
\label{sec:method}

LACA is a backbone-independent module for explicit temporal information
exchange between two hand-specific feature streams. In our principal
offline implementation, LACA replaces the same-index cross-hand fusion in
Polyphony \cite{zheng2026polyphony}, while retaining its remaining encoder
and segmentation components. LACA-C additionally enforces future-free
processing throughout the complete inference pipeline, as detailed in
Section~\ref{subsec:causal_variant}. As summarized in
Fig.~\ref{fig:laca_overview}(c), LACA estimates a directional distribution
over candidate temporal offsets, retrieves the corresponding cross-hand
information, and attenuates information transfer through an explicit null
state when no compatible dependency is identified.

\subsection{Problem Formulation}
\label{subsec:formulation}

Given an untrimmed video
$\mathbf{X}=(\mathbf{x}_1,\ldots,\mathbf{x}_N)$ containing $N$ native
frames, feature extraction and temporal sampling with stride $r$
produce hand-specific input sequences
$\mathbf{U}^{h}=[\mathbf{u}^{h}_1,\ldots,\mathbf{u}^{h}_T]$ on a
$T$-step sampled grid, where $h\in\{L,R\}$. In our Polyphony
implementation, $\mathbf{U}^{h}$ contains the hand-specific MAS
features. The ground-truth labels are sampled on the same grid, yielding
$\mathbf{Y}^{h}=(y^{h}_1,\ldots,y^{h}_T)$. A temporal encoder maps each
input stream to
\begin{equation}
    \mathbf{H}^{h}
    =
    [\mathbf{h}^{h}_1,\ldots,\mathbf{h}^{h}_T]
    \in\mathbb{R}^{T\times d},
    \qquad h\in\{L,R\}.
    \label{eq:hand_features}
\end{equation}
Here, $t$ indexes the sampled temporal grid rather than the native
camera-frame grid. Accordingly, $K$, $\sigma$, and all other
temporal-window quantities are measured in sampled time steps.

Let $\bar{h}$ denote the hand opposite to $h$. For each target feature $\mathbf{h}^{h}_t$, LACA searches the other-hand stream over the bounded offset set
\begin{equation}
    \Delta_K=\{-K,\ldots,-1,0,1,\ldots,K\}.
    \label{eq:offset_set}
\end{equation}
For the direction $h\leftarrow\bar{h}$, an offset $\delta<0$ retrieves an earlier feature from the other hand, whereas $\delta>0$ retrieves a later feature. Hence, the sign of $\delta$ explicitly represents the observed lead--lag direction.

\subsection{Bidirectional Lag-Aware Alignment}
\label{subsec:lag_alignment}

For each direction, the target-hand feature is projected to a query, while the candidate other-hand features are projected to keys and values:
\begin{equation}
\begin{aligned}
    \mathbf{q}^{h}_t
        &=\mathbf{W}^{h}_{Q}\mathbf{h}^{h}_t,\\
    \mathbf{k}^{\bar{h}}_{t+\delta}
        &=\mathbf{W}^{h}_{K}\mathbf{h}^{\bar{h}}_{t+\delta},\\
    \mathbf{v}^{\bar{h}}_{t+\delta}
        &=\mathbf{W}^{h}_{V}\mathbf{h}^{\bar{h}}_{t+\delta}.
\end{aligned}
\label{eq:qkv}
\end{equation}

The projection dimensions are
$\mathbf{W}_{Q}^{h},\mathbf{W}_{K}^{h},\mathbf{W}_{V}^{h}
\in\mathbb{R}^{d_a\times d}$,
$\mathbf{W}_{O}^{h}\in\mathbb{R}^{d\times d_a}$, and
$\mathbf{k}_{\varnothing}^{h}\in\mathbb{R}^{d_a}$.

Following scaled dot-product attention \cite{vaswani2017attention}, the compatibility score for each valid offset is
\begin{equation}
    e^{h\leftarrow\bar{h}}_{t,\delta}
    =
    \frac{
    (\mathbf{q}^{h}_t)^{\top}
    \mathbf{k}^{\bar{h}}_{t+\delta}}
    {\sqrt{d_a}},
    \qquad
    \delta\in\mathcal{V}_t,
    \label{eq:lag_score}
\end{equation}
where $d_a$ is the alignment dimension and
$\mathcal{V}_t=\{\delta\in\Delta_K:1\leq t+\delta\leq T\}$ excludes indices outside the video.

A learned null key $\mathbf{k}^{h}_{\varnothing}$ provides an additional score,
\begin{equation}
    e^{h\leftarrow\bar{h}}_{t,\varnothing}
    =
    \frac{
    (\mathbf{q}^{h}_t)^{\top}\mathbf{k}^{h}_{\varnothing}}
    {\sqrt{d_a}}.
    \label{eq:null_score}
\end{equation}
The lag and null scores are normalized jointly:
\begin{equation}
    \boldsymbol{\pi}^{h\leftarrow\bar{h}}_t
    =
    \operatorname{softmax}
    \left(
    \{e^{h\leftarrow\bar{h}}_{t,\delta}\}_{\delta\in\mathcal{V}_t},
    e^{h\leftarrow\bar{h}}_{t,\varnothing}
    \right).
    \label{eq:lag_distribution}
\end{equation}
The aligned cross-hand representation is computed only from the non-null probabilities:
\begin{equation}
    \widetilde{\mathbf{h}}^{\bar{h}\rightarrow h}_t
    =
    \sum_{\delta\in\mathcal{V}_t}
    \pi^{h\leftarrow\bar{h}}_{t,\delta}
    \mathbf{v}^{\bar{h}}_{t+\delta},
    \qquad
    \mathbf{z}^{h}_t
    =
    \mathbf{h}^{h}_t+
    \mathbf{W}^{h}_{O}
    \widetilde{\mathbf{h}}^{\bar{h}\rightarrow h}_t.
    \label{eq:aligned_fusion}
\end{equation}

In the Polyphony instantiation, $\mathbf{z}^{h}_t$ replaces the output
of the original same-index cross-hand fusion at the post-encoder fusion
site and is supplied as the conditioning representation to the
corresponding hand-specific denoising decoder. LACA is applied once at
this fusion site; the remaining encoder and decoder operations are
unchanged.

Because the non-null probabilities are not renormalized, their total mass
is $1-\pi^{h\leftarrow\bar h}_{t,\varnothing}$. The null probability
therefore gates the cross-hand update in
Eq.~\eqref{eq:aligned_fusion}. As $\pi^{h\leftarrow\bar h}_{t,\varnothing}\rightarrow1$, the transferred term vanishes and $\mathbf{z}_{t}^{h}\rightarrow\mathbf{h}_{t}^{h}$.
The same operation is applied independently in the $L\leftarrow R$ and $R\leftarrow L$ directions, allowing asymmetric
temporal relationships.


\subsection{Compatibility-Aware Transition Supervision}
\label{subsec:lag_supervision}

Temporal proximity alone does not establish that two transitions are related. We therefore construct lag targets using both transition semantics and temporal distance. From the frame-level training labels, a hand-specific transition is identified by
\begin{equation}
    b_t^h
    =
    \mathbb{I}[y_t^h\neq y_{t-1}^h],
    \qquad
    r_t^h=(y_{t-1}^h\rightarrow y_t^h),
    \label{eq:transition}
\end{equation}
where $r_t^h$ is the ordered transition type. We define the transition-anchor set for hand $h$ as
\begin{equation}
    \mathcal{T}^{h}
    =
    \{(t,r_t^h):b_t^h=1,\;t=2,\ldots,T\}.
    \label{eq:transition_anchors}
\end{equation}

Let $\mathcal{D}_{\mathrm{train}}$ denote the training videos used to
estimate transition compatibility, and let $\mathcal{T}^{h}_{v}$ denote
the transition-anchor set of hand $h$ in video $v$. For transition
types $a$ and $b$, we define
\begin{equation}
\begin{aligned}
N^{h\leftarrow\bar h}(a,b)
={}&
\sum_{v\in\mathcal{D}_{\mathrm{train}}}
\sum_{(t,a')\in\mathcal{T}^{h}_{v}}
\sum_{(\tau,b')\in\mathcal{T}^{\bar h}_{v}}
\\[-1mm]
&\mathbb{I}\!\left[
a'=a,\;b'=b,\;|\tau-t|\leq K
\right].
\end{aligned}
\label{eq:compatibility_count}
\end{equation}
Thus, every target--candidate transition pair occurring within the
$\pm K$ sampled-step window in the same video contributes once.
Directional compatibility is then estimated as
\begin{equation}
    C^{h\leftarrow\bar{h}}(a,b)
    =
    \frac{
    N^{h\leftarrow\bar{h}}(a,b)+\epsilon}
    {
    \sum_{b'\in\mathcal{R}^{\bar h}}
    N^{h\leftarrow\bar{h}}(a,b')
    +\epsilon|\mathcal{R}^{\bar h}|},
    \label{eq:compatibility}
\end{equation}
where $\mathcal{R}^{\bar h}$ is the set of other-hand transition types
and $\epsilon>0$ is a smoothing constant.

For a target transition at $t$, candidate other-hand transitions are scored as
\begin{equation}
    \tau^{*}
    =
    \argmax_{\tau:\,b_{\tau}^{\bar{h}}=1,\;|\tau-t|\leq K}
    \left[
    C^{h\leftarrow\bar{h}}
    (r_t^h,r_\tau^{\bar{h}})
    -
    \alpha\frac{|\tau-t|}{K}
    \right],
    \qquad
    \delta^{*}=\tau^{*}-t.
    \label{eq:transition_matching}
\end{equation}

Matching is directional and anchor-wise: each target-hand transition
independently selects its highest-scoring other-hand candidate. Thus, the
same other-hand transition may be selected by more than one target
anchor. This construction supervises cross-hand information retrieval
rather than defining a global one-to-one event assignment.

For a nonempty candidate set, let $s^{h,*}_t$ denote the maximum value
of the bracketed score in Eq.~\eqref{eq:transition_matching}. The match
is accepted if and only if $s^{h,*}_t\geq\theta$. An accepted match produces a Gaussian soft target centered at $\delta^{*}$:
\begin{equation}
    q^h_{t,\delta}
    =
    \frac{
    \exp[-(\delta-\delta^{*})^2/(2\sigma^2)]}
    {\sum_{\delta'\in\mathcal{V}_t}
    \exp[-(\delta'-\delta^{*})^2/(2\sigma^2)]},
    \qquad
    q^h_{t,\varnothing}=0.
    \label{eq:soft_lag_target}
\end{equation}
If no candidate is available or $s^{h,*}_t<\theta$, we assign
$q^h_{t,\varnothing}=1$ and $q^h_{t,\delta}=0$ for every
$\delta\in\mathcal{V}_t$. Thus, an unrelated transition is not forcibly aligned with its temporally nearest neighbor.


\subsection{Learning Objective}
\label{subsec:objective}

Training augments the original segmentation objective with lag supervision:
\begin{equation}
    \mathcal{L}
    =
    \mathcal{L}_{\mathrm{base}}
    +
    \lambda_{\ell}\mathcal{L}_{\mathrm{lag}},
    \label{eq:total_loss}
\end{equation}
where $\mathcal{L}_{\mathrm{base}}$ is the original Polyphony objective
\cite{zheng2026polyphony}, including its encoder, denoising-decoder,
temporal-smoothing, boundary, and adaptive hand-weighting terms.

The lag-supervision term is
\begin{equation}
    \mathcal{L}_{\mathrm{lag}}
    =
    -\frac{1}{|\mathcal{M}|}
    \sum_{(h,t)\in\mathcal{M}}
    \sum_{\delta\in\mathcal{V}_t\cup\{\varnothing\}}
    q^h_{t,\delta}
    \log\pi^{h\leftarrow\bar{h}}_{t,\delta},
    \label{eq:lag_loss}
\end{equation}
where $\mathcal{M}$ contains all target-hand transition anchors from
both alignment directions. Each anchor receives either a matched soft
lag target or a null target. No direct lag target is imposed on
non-transition frames; at those frames, the alignment distribution is
optimized only indirectly through $\mathcal{L}_{\mathrm{base}}$ and
the shared alignment parameters. Explicit null supervision therefore
applies to unmatched or incompatible transition anchors rather than
to every interval of independent hand activity. The compatibility matrices and training lag targets are constructed
exclusively from annotations in $\mathcal{D}_{\mathrm{train}}$ and are not required during inference. 

Test annotations are never used for compatibility estimation, target
construction, checkpoint selection, or hyperparameter tuning.

\subsection{Future-Free Variant}
\label{subsec:causal_variant}

For online inference, LACA-C restricts the candidate offsets to
\begin{equation}
    \Delta_K^{\mathrm{C}}=\{-K,\ldots,-1,0\},
    \qquad
    \mathcal{V}_t^{\mathrm{C}}
    =
    \{\delta\in\Delta_K^{\mathrm{C}}:1\leq t+\delta\leq t\}.
    \label{eq:causal_offsets}
\end{equation}
In LACA-C, $\mathcal{V}_t^{\mathrm{C}}$ replaces $\mathcal{V}_t$ in
both the alignment and lag-loss computations. During lag-target
construction, the candidate set in
Eq.~\eqref{eq:transition_matching} is restricted to other-hand
transitions satisfying $\tau-t\in\mathcal{V}_t^{\mathrm{C}}$.
Positive-offset transitions are therefore never considered. If no
nonpositive candidate is available or its best score is below
$\theta$, the anchor is assigned to the null state. The Gaussian target
in Eq.~\eqref{eq:soft_lag_target} is normalized over
$\mathcal{V}_t^{\mathrm{C}}$.

To make the complete inference pipeline future-free, LACA-C uses strict
prefix-wise inference. Let $n_t$ denote the native-frame index
corresponding to sampled position $t$. The MAS vector
$\mathbf{u}^{h}_t$ is generated from a right-aligned visual window
ending at $n_t$; no frame after $n_t$ is used, and only past-side
padding is permitted at the beginning of a video. Semantic conditioning
is likewise evaluated using only the observed prefix. At time $t$, the
temporal encoder, LACA-C module, hand-specific denoising decoder, and
any temporal post-processing receive only
$\mathbf{U}^{L}_{1:t}$ and $\mathbf{U}^{R}_{1:t}$, and only the
prediction at the current position is retained. Consequently,
\begin{equation}
    \widehat{y}^{h}_t
    =
    f_h\!\left(
    \mathbf{U}^{L}_{1:t},
    \mathbf{U}^{R}_{1:t}
    \right),
    \qquad h\in\{L,R\},
    \label{eq:future_free_prediction}
\end{equation}
and the prediction at $t$ cannot depend on any observation occurring
after $n_t$.

\section{Experiments and Results}
\label{sec:experiments}

\subsection{Datasets and Experimental Protocol}
\label{subsec:protocol}

\paragraph{Datasets.}
We use the official HA-ViD \cite{zheng2023havid} and ATTACH \cite{aganian2023attach} partitions adopted by Polyphony \cite{zheng2026polyphony}. The HA-ViD action-segmentation subset contains 609 untrimmed view-specific videos recorded across three camera views with 75 classes, split into 486 training and 123 test sequences. From the released training set, we hold out a trial-grouped validation subset of 96 sequences from 32 physical trials, keeping all views of each trial together. The remaining 390 sequences constitute $\mathcal{D}_{\mathrm{train}}$ and are used for parameter optimization, compatibility estimation, and lag-target construction. Validation is used for checkpoint and hyperparameter selection and Polyphony's adaptive hand weighting. Test results use the validation-selected checkpoint without subsequent refitting. ATTACH contains 378 untrimmed view-specific videos recorded across three camera views from 42 participants. Using the official participant split and the 24-class-per-hand mapping of the original 51 labels, 28, 4, and 10 participants form $\mathcal{D}_{\mathrm{train}}$, validation, and test sets, respectively. Validation supports model and hyperparameter selection. It does not
contribute to compatibility estimation or training-target construction;
the test set is reserved strictly for final evaluation.

\paragraph{Implementation.}
Offline LACA replaces only Polyphony's same-index cross-hand fusion; its
remaining architecture, optimization schedule, validation criterion, and
$\mathcal{L}_{\mathrm{base}}$ are retained. LACA-C and Polyphony-C follow
the complete future-free procedure in
Section~\ref{subsec:causal_variant}. The offline models use Polyphony's original MAS features and original
offline clip construction. For LACA-C,
Polyphony-C, and OnlineTAS, the feature at sampled position $t$ is
extracted from a right-aligned clip containing the 16 consecutive native
frames
$\{\mathbf{x}_{n_t-15},\ldots,\mathbf{x}_{n_t}\}$,
sampled at unit temporal stride. Thus, each clip contains 16 native frames and spans 0.50~s between its
first and last frame timestamps and contains no frame occurring after $n_t$.

At the beginning of a video, unavailable past frames are replaced by
replicating the first available frame until a 16-frame clip is obtained.
The ADH-ViT is initialized from the corresponding offline checkpoint and
fine-tuned on $\mathcal{D}_{\mathrm{train}}$ using the right-aligned clip
construction. The semantic-conditioning TCN is retrained using causal
padding and prefix-only inputs so that its output at position $t$ cannot
depend on features after $t$. The same future-free MAS features are used
for LACA-C, Polyphony-C, and OnlineTAS.

Each MAS vector comprises a 768-D motion feature, a $C$-D action-logit
feature, and a 384-D semantic feature, giving 1,227 dimensions on
HA-ViD ($C=75$) and 1,176 dimensions on ATTACH ($C=24$);
$d=d_a=64$.

The sampling stride is $r=4$ native frames, yielding a
7.5-Hz feature grid on both datasets. Unless explicitly stated otherwise, alignment-window quantities, including $K$, $\sigma$, and $\rho$, are expressed in sampled time steps. Streaming cue confirmation, matching, and availability delay in Section~\ref{subsec:online_results} are evaluated on the native 30-Hz frame timeline. 
We use $\epsilon=1.0$, $K=15$, $\alpha=0.30$,
$\theta=0.20$, $\sigma=2.0$, $\eta=0.50$, and $\lambda_{\ell}=0.20$. 

For OnlineTAS, we train two independent
hand-specific models using the same future-free MAS inputs, temporal
grid, split, and validation criterion. Its reported parameter count is
the sum of both models (3.20\,M).

All learned-model performance and throughput results are reported as
mean~$\pm$~standard deviation over seeds 0, 1, and 2. The deterministic annotation analysis in Section~\ref{subsec:lag_results} is reported once.
Availability delay is summarized within each seed by the median over all matched cues and then across seeds by the mean and standard deviation of the three seed-wise medians.

Experiments use an NVIDIA RTX A6000 with batch size one. Parameter counts include trainable segmentation-stage modules only, while throughput excludes video decoding and MAS extraction. Offline throughput includes temporal encoding, fusion, and diffusion decoding; future-free throughput measures sequential current-position predictions. After 100 warm-up iterations, 10,000 sampled test positions are timed.

\paragraph{Metrics.}
Following the standard temporal action-segmentation protocol
\cite{farha2019mstcn}, we report framewise accuracy (Acc.), segmental edit
score (Edit), and $\mathrm{F1}@\{10,25,50\}$ for each hand. Boundary
localization is measured by class-agnostic boundary F1 (B-F1). Predicted
and ground-truth label-change points are matched separately for each hand
within $0.5$~s using a one-to-one assignment that first maximizes the
number of matches and then minimizes their total temporal distance.
Unmatched predictions and ground-truth boundaries are false positives and
false negatives, respectively. On a feature grid of rate $f_d$, a pair is
eligible when $|\hat{t}-t|/f_d\leq0.5$~s. All metrics are reported per
hand; two-hand averages are used only where explicitly identified.

\subsection{Empirical Cross-Hand Lag Structure}
\label{subsec:lag_results}

We assess whether the training annotations exhibit more robust nonzero cross-hand matches than temporally shifted within-trial controls under the proposed compatibility-aware matching rule. To avoid using a trial to estimate the compatibility matrix by which that same
trial is evaluated, we use trial-level cross-fitting: for each physical trial, compatibility is estimated from all remaining training trials and then applied to the held-out trial. Because the synchronized camera views of a physical trial share the
same hand-action annotation, transition anchors are extracted once from the common label timeline; camera views are not counted as independent annotation sequences in this analysis. After counting each physical trial once, let
$\mathcal{D}_{\mathrm{train}}^{\mathrm{trial}}$ denote the set of unique
training trials and define
\[
\mathcal{A}
=
\left\{
(v,h,t):
v\in\mathcal{D}_{\mathrm{train}}^{\mathrm{trial}},
\ h\in\{L,R\},
\ (t,r_{v,t}^{h})\in\mathcal{T}_{v}^{h}
\right\}.
\]
The robust-nonzero anchor rate is
\begin{equation}
R_{\mathrm{RNZ}}
=
\frac{
\sum_{(v,h,t)\in\mathcal{A}}
\mathbb{I}\!\left[
m_{v,t}^{h}=1
\land
|\delta_{v,t}^{h,\star}|>\rho
\right]
}{
|\mathcal{A}|
},
\label{eq:rnz_rate}
\end{equation}
where $m_{v,t}^{h}$ indicates an accepted compatibility-aware match and
$\rho=2$ sampled steps. The observed all-anchor rates are 44.7\% on
HA-ViD and 48.9\% on ATTACH. Among accepted robust-nonzero matches,
the median absolute lags are 5 and 6 sampled steps, respectively.

For each held-out physical trial $v$, let
$C_{-v}^{h\leftarrow\bar h}$ denote the compatibility matrix estimated
from all remaining training trials. Both the unshifted annotation
timeline of trial $v$ and every temporally shifted control derived from
that trial are evaluated using this same fixed matrix. For each of
$N_{\mathrm{perm}}=1000$ permutations, an admissible circular shift
$u_v$ is drawn independently for every trial, subject to
$\min(u_v,T_v-u_v)>K$, where $T_v$ is the number of sampled positions in trial $v$, and the other-hand transition timeline is shifted
by $u_v$. The pooled robust-nonzero all-anchor rate is then recomputed
across trials. The observed rates are 44.7\% on HA-ViD and 48.9\% on
ATTACH, whereas the mean shifted-control rates are 18.6\% and 21.3\%,
respectively. This descriptive separation is consistent with structured
cross-hand temporal organization relative to the temporal-shift control;
it does not establish causal influence between the hands.

\begin{figure*}[t]
    \centering
    \includegraphics[width=\textwidth]{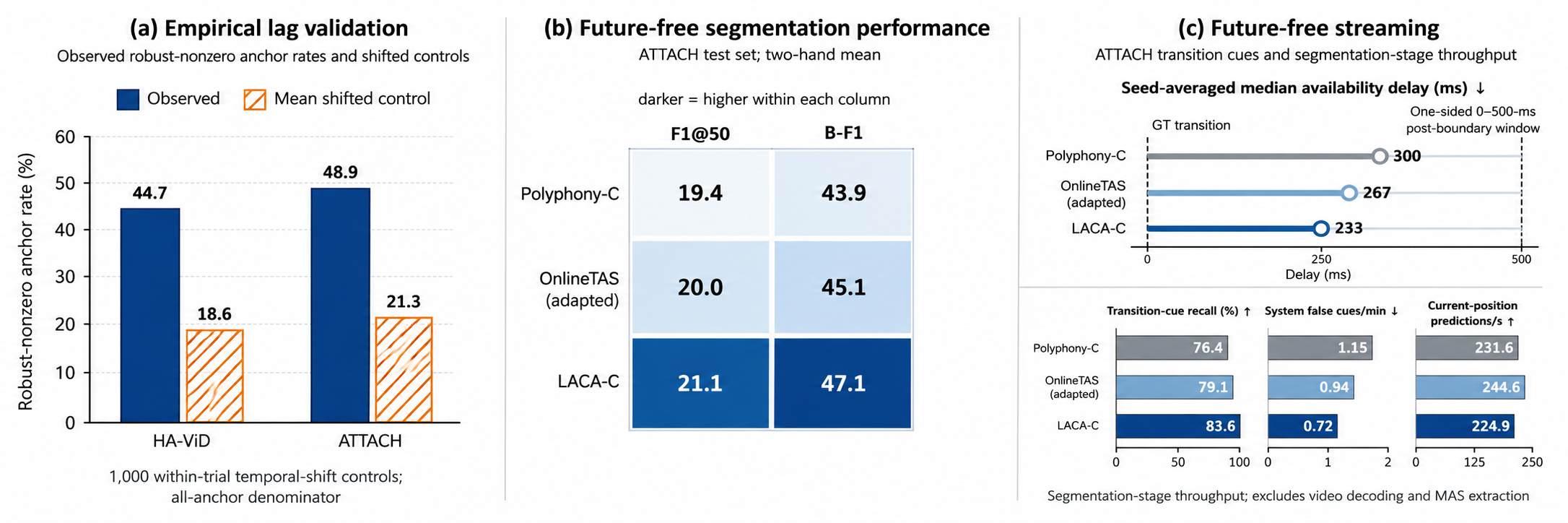}
   \caption{Empirical lag validation and future-free evaluation.
(a) Observed robust-nonzero transition-anchor rates on HA-ViD and ATTACH compared with the mean rates obtained from 1,000 within-trial temporal-shift controls, using the all-anchor denominator in Eq.~(18).
(b) Two-hand mean F1@50 and boundary F1 (B-F1) on the ATTACH test set for Polyphony-C, adapted OnlineTAS, and LACA-C under the same future-free protocol; darker cells indicate higher performance within each metric.
(c) Future-free streaming evaluation on ATTACH. The upper panel reports the seed-averaged median cue-availability delay under a one-sided 0--500-ms post-boundary matching window. The lower panels report transition-cue recall, system false cues per minute, and current-position prediction throughput. Throughput is measured for the segmentation stage and excludes video decoding and MAS feature extraction. Arrows indicate the preferred direction for each metric.}
    \label{fig:result_analysis}
\end{figure*}

\subsection{Segmentation and Boundary Localization}
\label{subsec:main_results}

Table~\ref{tab:main_results} compares LACA with recent dual-hand
action-segmentation methods and our controlled reproduction of
Polyphony. The reproduced results differ from the published Polyphony
values by at most 0.2 points, supporting the consistency of the
evaluation pipeline. On HA-ViD, LACA improves the two-hand mean
F1@50 from 40.4 to 42.5 and B-F1 from 56.5 to 59.6, corresponding
to gains of 2.1 and 3.1 points, respectively. On ATTACH, the
corresponding means increase from 19.9 to 21.8 for F1@50 and from
44.7 to 47.9 for B-F1, yielding gains of 1.9 and 3.2 points.
The hand-specific gains on HA-ViD and the larger B-F1 improvements on both datasets are consistent with the boundary-focused motivation of lag-aware retrieval.

\begin{table*}[t]
\centering
\caption{Comparison with dual-hand action-segmentation methods.
F1 reports F1@10/25/50. Values marked with $\dagger$ are taken from
Polyphony~\cite{zheng2026polyphony}, which did not report B-F1.
For HA-ViD left-hand F1@10, we use 58.8 from Polyphony's complete
supplementary/component results; its selected main-paper table prints
58.2. Values marked with $\ddagger$ are mean~$\pm$~SD over three
independently trained seeds. B-F1 follows the one-to-one
boundary-matching protocol in Section~\ref{subsec:protocol}.}
\label{tab:main_results}
\setlength{\tabcolsep}{3.0pt}
\resizebox{\textwidth}{!}{%
\begin{tabular}{lllcccccccc}
\toprule
& & &
\multicolumn{4}{c}{Left hand} &
\multicolumn{4}{c}{Right hand} \\
\cmidrule(lr){4-7}
\cmidrule(lr){8-11}
Dataset & Method & Input &
Acc. & Edit & F1@10/25/50 & B-F1 &
Acc. & Edit & F1@10/25/50 & B-F1 \\
\midrule

HA-ViD
& DiffAct~\cite{liu2023diffusion}$^\dagger$ & I3D
& 43.7 & 42.5 & 44.1/38.1/26.0 & --
& 44.4 & 47.9 & 48.8/42.4/28.4 & -- \\

& FACT~\cite{lu2024fact}$^\dagger$ & I3D
& 45.1 & 47.2 & 50.1/43.0/29.6 & --
& 43.8 & 45.3 & 47.6/40.3/27.1 & -- \\

& Polyphony$^\dagger$ & MAS
& 57.1 & 53.7 & 58.8/53.0/39.6 & --
& 60.6 & 54.8 & 61.5/55.1/41.3 & -- \\

& Polyphony (reproduced)$^\ddagger$ & MAS
& \mstd{57.0}{0.3}
& \mstd{53.6}{0.4}
& \mstd{58.6}{0.4}/\mstd{52.9}{0.5}/\mstd{39.5}{0.4}
& \mstd{55.2}{0.5}
& \mstd{60.5}{0.3}
& \mstd{54.7}{0.4}
& \mstd{61.4}{0.3}/\mstd{55.0}{0.4}/\mstd{41.2}{0.4}
& \mstd{57.8}{0.5} \\

& \textbf{LACA}$^\ddagger$ & MAS
& \textbf{\mstd{58.3}{0.3}}
& \textbf{\mstd{55.2}{0.4}}
& \textbf{\mstd{60.2}{0.4}/\mstd{54.7}{0.4}/\mstd{41.7}{0.3}}
& \textbf{\mstd{58.4}{0.4}}
& \textbf{\mstd{61.7}{0.2}}
& \textbf{\mstd{56.2}{0.3}}
& \textbf{\mstd{62.9}{0.3}/\mstd{56.8}{0.4}/\mstd{43.3}{0.3}}
& \textbf{\mstd{60.8}{0.4}} \\

\midrule

ATTACH
& DiffAct~\cite{liu2023diffusion}$^\dagger$ & I3D
& 47.5 & 44.1 & 40.7/31.4/15.3 & --
& 42.5 & 46.9 & 43.5/33.9/17.5 & -- \\

& FACT~\cite{lu2024fact}$^\dagger$ & I3D
& 45.8 & 46.8 & 39.1/29.6/14.3 & --
& 40.1 & 46.4 & 41.0/31.3/15.6 & -- \\

& Polyphony$^\dagger$ & MAS
& 52.8 & 47.8 & 45.7/36.9/19.2 & --
& 47.3 & 49.7 & 46.9/37.7/20.8 & -- \\

& Polyphony (reproduced)$^\ddagger$ & MAS
& \mstd{52.7}{0.4}
& \mstd{47.7}{0.5}
& \mstd{45.6}{0.4}/\mstd{36.8}{0.5}/\mstd{19.1}{0.3}
& \mstd{43.8}{0.5}
& \mstd{47.2}{0.4}
& \mstd{49.6}{0.4}
& \mstd{46.8}{0.4}/\mstd{37.6}{0.4}/\mstd{20.7}{0.3}
& \mstd{45.5}{0.5} \\

& \textbf{LACA}$^\ddagger$ & MAS
& \textbf{\mstd{54.0}{0.3}}
& \textbf{\mstd{49.2}{0.4}}
& \textbf{\mstd{47.2}{0.4}/\mstd{38.5}{0.4}/\mstd{20.9}{0.3}}
& \textbf{\mstd{47.0}{0.5}}
& \textbf{\mstd{48.5}{0.3}}
& \textbf{\mstd{51.1}{0.4}}
& \textbf{\mstd{48.4}{0.4}/\mstd{39.4}{0.4}/\mstd{22.7}{0.3}}
& \textbf{\mstd{48.8}{0.5}} \\

\bottomrule
\end{tabular}}
\end{table*}

\subsection{Ablation and Efficiency}
\label{subsec:ablation}

Table~\ref{tab:ablation_online} evaluates the contributions of the
alignment components on ATTACH. Local temporal attention searches the
same offset window $\Delta_K$ as LACA but uses neither lag supervision
nor an explicit null state. The variant without
$\mathcal{L}_{\mathrm{lag}}$ retains the null state but learns the
alignment only through $\mathcal{L}_{\mathrm{base}}$. The variant
without the null state removes $\mathbf{k}_{\varnothing}^{h}$, normalizes over
non-null offsets, and excludes unmatched transition anchors from
$\mathcal{L}_{\mathrm{lag}}$.

Local temporal attention improves the two-hand mean F1@50 by only
0.6 points over same-time fusion. Removing lag supervision reduces
F1@50 by 0.9 points, whereas removing the null state reduces B-F1 by
1.9 points. These results show that the gain cannot be attributed
solely to a larger temporal search window: both explicit lag
supervision and suppression of unsupported cross-hand transfer are
important. Replacing the original same-index fusion with full LACA increases the number of active trainable parameters by approximately
0.0086\,M under $d=d_a=64$. Segmentation-stage throughput decreases
from 238.4 to 211.7 predictions per second.

\begin{table*}[t]
\centering
\caption{Ablation, efficiency, and future-free evaluation on ATTACH.
F1@50 and B-F1 are two-hand means reported as mean~$\pm$~SD over
three independently trained seeds. Active parameter counts include only
trainable segmentation-stage modules used in the forward pass and
exclude feature extraction. Throughput is reported as sampled
predictions per second and excludes video decoding and MAS extraction.
Polyphony-C uses the same future-free input protocol without lag-aware
retrieval. OnlineTAS~\cite{zhong2024onlinetas} is adapted to dual-hand
labels and evaluated using the same split. System FPM is the total
number of unmatched cues from both hands divided by physical video
duration in minutes.}
\label{tab:ablation_online}
\setlength{\tabcolsep}{3.5pt}
\resizebox{\textwidth}{!}{%
\begin{tabular}{lccccccc}
\toprule
Variant &
F1@50 $\uparrow$ &
B-F1 $\uparrow$ &
Active params (M) $\downarrow$ &
Predictions/s $\uparrow$ &
Recall (\%) $\uparrow$ &
Availability delay (ms) $\downarrow$ &
System FPM $\downarrow$ \\
\midrule

Polyphony (same-time fusion)
& \mstd{19.9}{0.3}
& \mstd{44.7}{0.5}
& 18.760
& \mstd{238.4}{2.1}
& -- & -- & -- \\

Local temporal attention
& \mstd{20.5}{0.4}
& \mstd{45.2}{0.6}
& 18.768
& \mstd{213.5}{1.8}
& -- & -- & -- \\

LACA w/o $\mathcal{L}_{\mathrm{lag}}$
& \mstd{20.9}{0.3}
& \mstd{46.2}{0.5}
& 18.769
& \mstd{211.8}{1.7}
& -- & -- & -- \\

LACA w/o explicit null state
& \mstd{21.2}{0.4}
& \mstd{46.0}{0.6}
& 18.768
& \mstd{213.0}{1.9}
& -- & -- & -- \\

\textbf{LACA}
& \textbf{\mstd{21.8}{0.3}}
& \textbf{\mstd{47.9}{0.4}}
& 18.769
& \mstd{211.7}{1.6}
& -- & -- & -- \\

\midrule

Polyphony-C (future-free)
& \mstd{19.4}{0.4}
& \mstd{43.9}{0.6}
& 18.760
& \mstd{231.6}{2.3}
& \mstd{76.4}{0.9}
& \mstd{300}{17}
& \mstd{1.15}{0.06} \\

OnlineTAS (adapted)
& \mstd{20.0}{0.3}
& \mstd{45.1}{0.5}
& 3.200
& \textbf{\mstd{244.6}{2.0}}
& \mstd{79.1}{0.8}
& \mstd{267}{14}
& \mstd{0.94}{0.05} \\

\textbf{LACA-C}
& \textbf{\mstd{21.1}{0.3}}
& \textbf{\mstd{47.1}{0.5}}
& 18.769
& \mstd{224.9}{1.8}
& \textbf{\mstd{83.6}{0.7}}
& \textbf{\mstd{233}{12}}
& \textbf{\mstd{0.72}{0.04}} \\

\bottomrule
\end{tabular}}
\end{table*}

\subsection{Future-Free Streaming Case Study}
\label{subsec:online_results}

We evaluate LACA-C sequentially on ATTACH using only observations available up to the current sampled position. Predictions are expanded to the native 30-Hz timeline by holding each sampled output until the next position. A transition cue is emitted after the new predicted state remains unchanged for $s=5$ native frames and is typed by its ordered before--after label pair. Cues are matched separately per video and hand to same-type ground-truth transitions within a one-sided 0--0.5-s window, using one-to-one assignment that maximizes matches and then minimizes delay. Recall is micro-averaged across both hands, system FPM counts unmatched cues per physical video minute, and availability delay is the median over matched cues.

As shown in Table~\ref{tab:ablation_online}, LACA-C achieves 83.6\%
transition recall, a seed-averaged median delay of 233~ms, and
0.72~FPM at 224.9~Predictions/s. Its two-hand mean F1@50 is 21.1, which is 0.7 points below offline LACA (21.8); the LACA-C result is obtained under the complete future-free pipeline. This comparison jointly
reflects right-aligned feature construction, restriction to nonpositive
cross-hand offsets, and sequential prefix-wise inference, rather than the effect of any one component in isolation. Figure~\ref{fig:result_analysis}(c)
summarizes availability delay, transition recall, system false-cue
rate, and segmentation-stage throughput.

\section{Conclusion and Future Work}
\label{sec:conclusion}

This work addressed temporal asynchrony in dual-hand action segmentation through Lag-Aware Cross-Hand Alignment (LACA), a lightweight module that replaces same-index inter-hand fusion with explicit, directional, and time-varying alignment over candidate offsets. A learned null state suppresses unsupported cross-hand transfer, while compatibility-aware targets are derived directly from frame-level training annotations. Analyses of HA-ViD and ATTACH showed substantially higher robust-nonzero transition-anchor rates than temporally shifted controls, indicating structured cross-hand temporal relationships without implying causal influence.

Integrated into the reproduced Polyphony framework, LACA consistently improved segmentation and boundary localization on both datasets, with particularly clear B-F1 gains and only a small parameter increase. Ablation results confirmed that the improvements arise not merely from a larger temporal search window, but from both explicit lag supervision and null-state gating. The future-free LACA-C variant also improved segmentation performance and transition-cue recall while reducing availability delay and false-cue rate relative to future-free baselines using only current and past observations.

The study is limited to two assembly-oriented datasets, with future-free evaluation conducted only on ATTACH. It also relies on frame-level transition labels and a fixed search range, while throughput excludes feature extraction. Future work will examine broader egocentric settings, adaptive or weakly supervised alignment, object-aware cues, efficient incremental decoding, and end-to-end real-time evaluation.

\bibliographystyle{elsarticle-num}
\bibliography{References}

\end{document}